# A convoy of magnetic millirobots transports endoscopic instruments for minimally-invasive surgery

*Moonkwang Jeong[+], Xiangzhou Tan[+], Felix Fischer and Tian Qiu\**

M. Jeong

Cyber Valley group - Biomedical Microsystems, Institute of Physical Chemistry, University of Stuttgart, Pfaffenwaldring 55, 70569 Stuttgart, Germany

X. Tan

Department of General Surgery, Xiangya Hospital, Central South University, Changsha 410008, China

International Joint Research Center of Minimally Invasive Endoscopic Technology Equipment & Standards, Changsha 410008, China

F. Fischer

Division of Smart Technologies for Tumor Therapy, German Cancer Research Center (DKFZ) Site Dresden, Blasewitzer Str. 80, 01307 Dresden, Germany

Faculty of Engineering Sciences, University of Heidelberg, Heidelberg, Germany

T. Qiu

Division of Smart Technologies for Tumor Therapy, German Cancer Research Center (DKFZ) Site Dresden, Blasewitzer Str. 80, 01307 Dresden, Germany

Faculty of Medicine Carl Gustav Carus, Technical University Dresden, Germany

Faculty of Electrical and Computer Engineering, Technical University Dresden, Germany

+ These authors contribute equally to this work.

\* Corresponding author, e-mail: tian.qiu@dkfz.de

((**Abstract** text. Maximum length 200 words. Written in the present tense.))

Small-scale robots offer significant potential in minimally-invasive medical procedures. Due to the nature of soft biological tissues, however, robots are exposed to complex environments with various challenges in locomotion, which is essential to overcome for useful medical tasks. A single mini-robot often provides insufficient force on slippery biological surfaces to carry medical instruments, such as a fluid catheter or an electrical wire. Here, for the first time, we report a team of millirobots (TrainBot) that can generate around two times higher actuating force than a TrainBot unit by forming a convoy to collaboratively carry long and heavy cargos. The feet of each unit are optimized to increase the propulsive force around three times so that it can effectively crawl on slippery biological surfaces. A human-scale permanent magnetic set-up is developed to wirelessly actuate and control the TrainBot to transport heavy and lengthy loads through narrow biological lumens, such as the intestine and the bile duct. We demonstrate the first electrocauterization performed by the TrainBot to relieve a biliary obstruction and open a tunnel for fluid drainage and drug delivery. The developed technology sheds light on the collaborative strategy of small-scale robots for future minimally-invasive surgical procedures.

1. Introduction

Small-scale robots have shown their great potential for biomedical applications, for example, for targeted drug delivery, sensing and microsurgery.[1–3] However, a single robot at a small scale often generates insufficient force to perform useful minimally-invasive surgical procedures, for instance, to deploy a long and thin catheter for drainage or drug delivery, or to resect soft tissues and open up a stricture. Minimally-invasive medical procedures often require a long tube for fluid transportation, *e.g.*, the drainage of biological fluids and the administration of drugs; or an electric electrode connected with wires for electrocautery procedures. These tubes or wires are often in the size range of 2-13 mm in diameter and over half a meter long, which are too long and too heavy to be carried by a TrainBot unit. To the best of our knowledge, it has not been shown that a team of millimeter-scale wireless robots can cooperate and transport an endoscopic instrument to perform a minimally-invasive surgical task.

Robots designed and developed in different sizes from nano- to centimeter-scale for surgical tasks have been reported including a nano-scale robot that can penetrate the vitreous body of the eye for drug delivery,[4] a tadpole-like[5] and a snake-like magnetic micrometer-scale robots swarm[6] that can swim in a fluid, a millimeter-scale soft catheter for endovascular surgery[7] and for minimally invasive bioprinting[8], a centimeter-scale tethered robot for navigating in a silicone-based phantom[9] and a porcine colon (*ex vivo*)[10]. Recent advances have shown small-scale robots that can overcome many challenges existing in real biological environments. For example, in the human body, robots need to crawl on a slippery surface of soft tissue covered by mucus.[11,12] It is a much more challenging task, as the friction is not large enough to support the forward motion and it often leads to a slip motion[13,14] that restricts the overall actuating force of the small-scale robots. Previous studies show that surface chemistry increases friction [13] and multiple legs help a robot to maintain its position.[13,15,16] In addition to that, crawling is a widely adopted mechanism for millimeter-scale wireless robots to locomote on biological tissues[13,17,18], on a hard substrate [19] and in a microchannel filled with silicone oil [20]. However, optimization of the robots' legs or feet has not been performed to generate higher actuating force as well as a grouping of the single robots. Second, biological soft tissues covered with a layer of interfacial fluid often exhibit slippery surfaces because the layer acts as a lubricant, which is a major obstacle for tissue adhesion.[12] For instance, the mucus layer in rats is observed up to 940 μm in thickness.[21] Moreover, it has not been reported that a team of millirobots can work synergistically to transport endoscopic instruments. Here, in this paper, we develop a convoy of millirobots (TrainBot) with optimized feet design to firmly anchor to the slippery biological surface and provide an individual driving force with distributed contact points on the surface, thus the team of robots offers a larger overall forward propulsive force and a lower chance of being stuck at the slippery terrain. Lastly, it is difficult to generate sufficient magnetic field strength for the wireless actuation of the robots, while maintaining a large enough space to accommodate a human patient, as the magnetic field decays significantly over distance.[22] Manipulation of small-scale robots is often performed using electromagnetic coil set-ups, [4,6,13,19,20,23–28] which have high controllability but very limited working space.[29,30] In contrast, permanent magnet arrays [17,31–34] have larger accessible volumes but are more difficult to design and control.[17,29,30,35] Recently, we reported an actuation set-up for the wireless actuation of a single robot using an oscillating magnetic field to walk on a slippery surface.[17] In this work, we improved the magnetic actuation set-up based on an array of three permanent magnets. Our magnetic actuation setup does not require a complex control mechanism and can simultaneously actuate multiple robots in a relatively gradient-free large working volume. This capability enables precise control of crawling using a rotating magnetic field within a volume that could potentially accommodate a human patient in the future. Cholangiocarcinoma is one of the malignant tumors originating from the bile duct, with unacceptable high morbidity and mortality.[36] Bile duct obstruction is a common symptom for those patients with cholangiocarcinoma, affecting around 30-50% of patients.[37,38] The obstruction in the bile duct blocks the transportation of the bile juice from the liver to the gallbladder and the small intestine, which could lead to serious complications, such as jaundice, cholangitis, multiple organ failure, etc. Endoscopic retrograde cholangiopancreatography (ERCP) is considered one of the first-line procedures for the diagnosis of a biliary obstruction.[39] A commercial clinical duodenoscope is normally used during the ERCP procedures. The duodenoscope for single-use has 11.3-13.7 mm outer diameters and 1240 mm working length with a working channel of 4.2 mm in diameter.[40] It is inserted through the upper digestive tract to the duodenum, and then into the bile duct through the Ampulla of Vater (as illustrated in **Figure 1a**). However, the procedure is very difficult, if not impossible, relying on the current endoscopic instrumentation to reach the upper bile ducts. The main difficulty of the procedure is the sharp physiological angle between the bile duct and the duodenum. Retrogradely pushing a flexible instrument, *e.g.* a catheter or an electric wire, through the bile duct is not feasible and may lead to the danger of tissue perforation.[41,42] Therefore, it will be of great benefit to the surgical procedure, if an instrument can actively move instead of being passively pushed through the bile duct. Here, for the first time, we report a convoy of wirelessly-driven magnetic millirobots (TrainBot) that can transport endoscopic instruments, such as a cannula or an electrocautery tool, for minimally-invasive surgical procedures. Each TrainBot unit is $1.6 \times 2 \times 3.1$ mm$^3$ in size, and a convoy of three TrainBot units can effectively carry an endoscopic instrument that is 250 mm in length and 70 mg in weight to crawl on the slippery and rough biological surfaces of the intestine and the bile duct. The TrainBot unit's feet are optimized to anchor to the slippery interfaces to enable effective locomotion and to follow complex trajectories. We show that each TrainBot unit is actuated simultaneously by a rotating magnetic field, and generates higher propulsive force due to the increase of the feet-surface contact area. The convoy of the units is able to controllably navigate in an *ex vivo* porcine bile duct, and transport an electrocautery tool to perform the first electrocautery surgery by the TrainBot. Our work sheds light on using a team of millirobots to collaboratively transport an endoscopic instrument through biological lumens for minimally-invasive surgical procedures.

2. Results and Discussion
2.1. TrainBot for endoscopic surgery in the bile duct
**Figure 1a** illustrates the endoscopic procedure enabled by the TrainBot, i.e., a group of TrainBot units connected via a flexible instrument. An endoscope passes through the upper GI tract and reaches the duodenum. The TrainBot is then inserted by pushing it through the endoscope's tool channel into the bile duct via the Ampulla of Vater. Once inserted into the bile duct, three units are simultaneously actuated by an external rotating magnetic field. The TrainBot crawls forward and, in the meantime, transports the long surgical instrument, *e.g.* an electric wire, to the target location, *i.e.*, a blockage at the upper bile duct. The electrocautery device is activated to resect the blocked soft tissue to create an opening for the subsequent process. After the retrieval of the electrocautery instrument, TrainBot carries another instrument, *e.g.,* a catheter, again to the target location for biliary juice drainage or local drug delivery. **Figure 1b** shows the enlarged view of the second TrainBot unit and **Figure 1c** shows the rotational magnetic field (**B**) for the actuation and the corresponding moving direction (*v*). The magnetic field **B** is generated by an array of permanent magnets, which consists of a pair of static magnets and a rotating magnet (see details in the Section below and **Figure 2**).

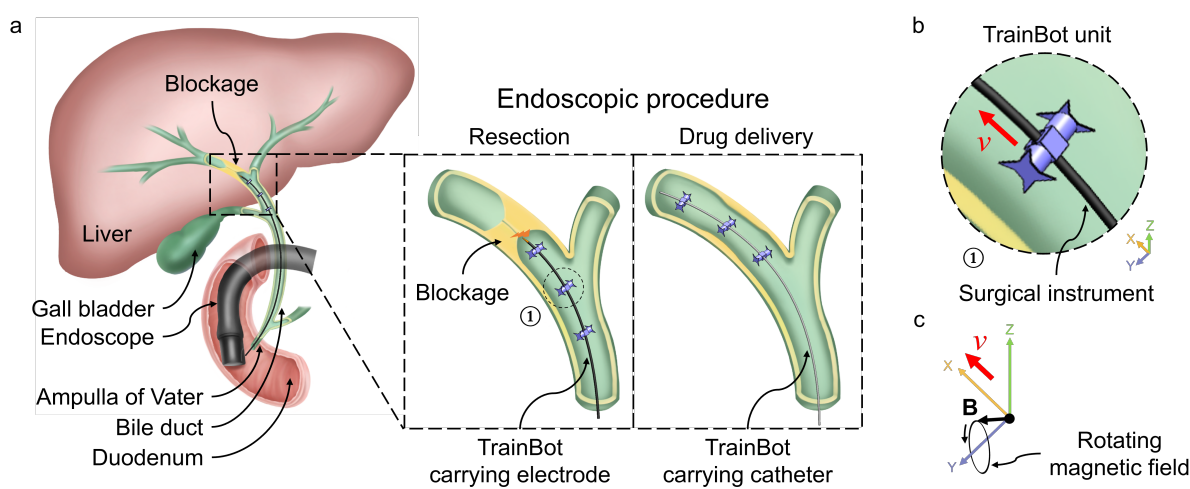

**Figure 1.** Schematic illustration of a TrainBot carrying a surgical instrument through the bile duct for an endoscopic procedure. (a) Insertion of the TrainBot via the tool channel of an endoscope, which reaches the duodenum through the upper GI tract. In this case, the TrainBot consists of three units articulated on a surgical instrument and carries the instrument towards the obstruction. An electrocautery tool is transported to open up the obstruction, then a catheter is transported for fluid drainage and drug delivery. (b) Enlarged view of a single TrainBot unit. (c) External rotating magnetic field **B** which drives and controls the TrainBot.

2.2. Magnetic actuation and crawling of a single TrainBot unit

A TrainBot unit is actuated wirelessly with a customized magnetic actuation set-up, as shown in **Figure 2a**. The set-up consists of a pair of two static magnets and one rotating magnet and generates a rotating magnetic field on the surface of a virtual cone over a full period (details in **Figure 3**). The space among the three magnets is defined as the accessible volume (details in **Figure S1**), which is ⌀ 50 × 28 cm$^2$, large enough to accommodate a human patient in a prone position. The central volume in the space is defined as the working space, where the TrainBot can be actuated, which is around 45 × 45 × 45 mm$^3$ in size (where the substrate is shown in **Figure 2a**). The detailed design of a TrainBot unit is shown in the enlarged view. It comprises three components: two actuators (colored in red), one coupler (yellow), and two feet with spikes (green). Two cylindrical permanent magnets (actuators) are attached on each side of a coupler with the magnetic moments aligned in the axial direction. Two non-magnetic feet, fabricated using Molybdenum due to good biocompatibility and high Young's modulus, are attached to the two sides of the actuators. Each side is equipped with four spikes to anchor to biological surfaces in all directions for navigation in the complex geometry of the bile duct. A through-hole is designed in the coupler for the insertion of the instrument and the connection of multiple units (as shown in **Figure 1a**). The actuators and coupler are non-biocompatible but feet are biocompatible. One possible way for TrainBot to be biocompatible is a thin-layer coating using a biocompatible polymer excluding the feet for better anchoring. However, the TrainBot is designed to be retrieved after a surgical procedure.

The motion of the TrainBot unit is controlled by the magnetic field **B**, defined by three independent parameters: $\alpha$, $\beta$, and $\varphi$ (**Figure 2a**). By changing the angle $\alpha$ of the rotating magnet, the unit moves either forward or backward. The angle $\beta$ controls the crawling direction of the robot. The angle $\varphi$ is defined as the pose angle of the robot, which is determined by the distance between the TrainBot unit and the rotating magnet.

The crawling of a TrainBot unit is shown in a sequence in **Figures 2b** and **2c** (also see **Video S1**). Half of the full cycle of the crawling ($\alpha = 90°$ to $270°$) is split into steps every $30°$ to illustrate the motion. The experimental results, shown on the right side, align perfectly with the simulation in both the front view and the top view, respectively. The rotation of the TrainBot unit follows the external rotating magnetic field. During a full cycle, one-foot anchors to a substrate while lifting the other foot, and the TrainBot unit rotates its body with the rotating magnetic field that results in a stride length either forward or backward. The direction of the unit's magnetic moment is marked as the blue arrow on the right side in the simulation results. The maximum angle $\varphi_{max}$ is defined as the TrainBot unit's maximal pose angle when $\alpha = 180°$ as shown in the fourth image. $\varphi_{max}$ can be adjusted by reconfiguring the magnetic actuation set-up. In the shown experimental setting, the static magnetic field has a flux density of 4.7 mT and the rotating field has a maximal magnitude of 3.6 mT. Thus, it results in a maximal pose angle of 37°, derived using **Equations 1-4**. In the second half of the cycle, the TrainBot unit alternates the two feet for anchoring and moving to take another step, which is characterized in detail in the next section.

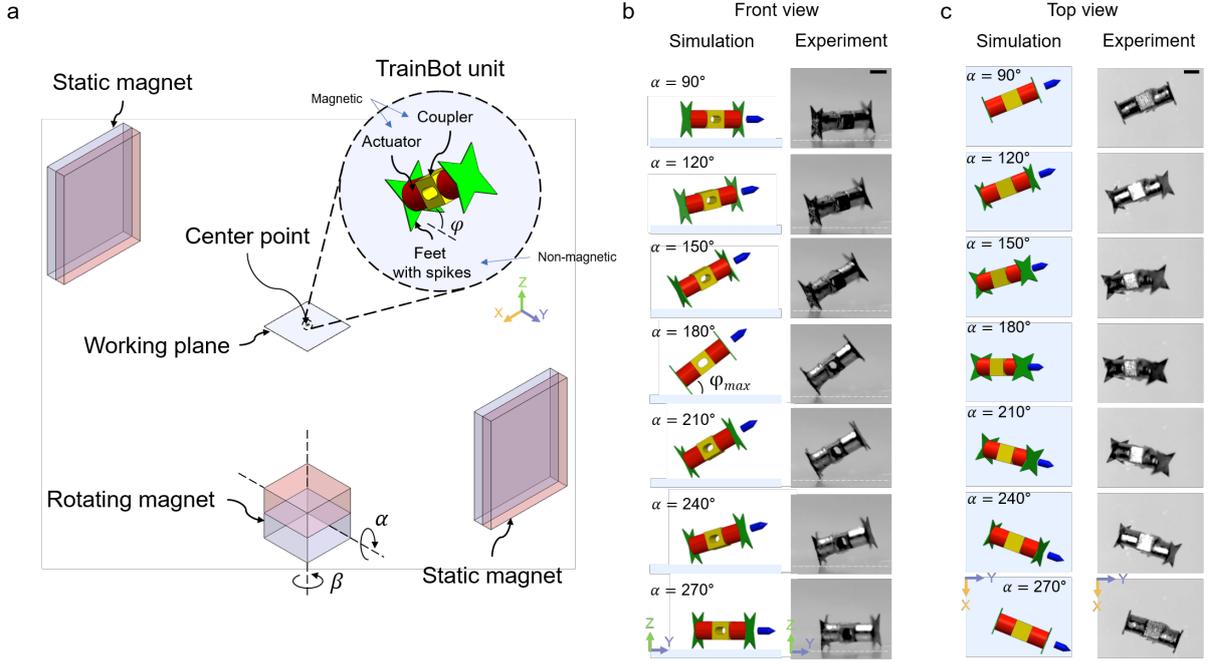

**Figure 2.** The crawling of a TrainBot unit wirelessly driven by a magnetic actuation set-up. (a) Schematic of the magnetic actuation set-up, which consists of three magnets (two static and one rotating magnet). Three independent parameters ($\alpha$, $\beta$, and $\varphi$) control the locomotion of the TrainBot unit. The center point of the workspace, where three center lines of each magnet are crossing, is defined as (0, 0, 0). TrainBot unit consists of three parts: an actuator, a coupler, and feet. (b-c) Simulation and experimental results of the crawling in a time series ($\alpha$ = 90°-270°, shown as front-view and top-view, respectively). All scale bars are 1 mm.

A superimposed magnetic field leads to the crawling of the TrainBot unit. The magnetic field is modeled as an ideal magnetic dipole, which generates a magnetic field at a given point.[43]

$$\mathbf{B} = \frac{\mu_0}{4\pi} \frac{3\hat{\mathbf{r}}(\hat{\mathbf{r}} \cdot \mathbf{m}) - \mathbf{m}}{|\mathbf{r}|^3}, \qquad (1)$$

where **r** is the vector from the center of the magnetic dipole to the measurement location (the center point of the set-up), **m** is the magnetic dipole moment and $\mu_0$ is the vacuum permeability. The pair of static magnets dominates $B_y$ and the rotating magnet dominates $B_x$ and $B_z$. The magnetic fields in all axes are derived as follows

$$B_x = -\frac{\mu_0 m_r}{4\pi h^3} \sin\alpha, \qquad (2)$$

$$B_y = \frac{\mu_0 m_s}{\pi w^3}, \qquad (3)$$

$$B_z = -\frac{\mu_0 m_r}{2\pi h^3}\cos\alpha \qquad (4)$$

where *h* and *w* stand for the distance from the center of static magnets and the rotating magnet to the measurement location (center point in **Figure 3a**), respectively. $m_s$ and $m_r$ are the magnetic dipole moments of the static and the rotating magnet, respectively, which are calculated based on the remanence and the volume of the magnets specified in the datasheets (see details in Experimental Section). $\alpha$ is the rotation angle of the rotating magnet. The magnetic field **B** during a cycle is measured and simulated at the center point shown in **Figure 3a**. As shown in **Figure 3b** and **Video S2**, the measured results fit the analytical solutions very well. In the analytical solution, *h* and *w* are fixed at 190.5 mm and 259.75 mm (center-to-center distance), respectively. It shows that the static magnet pair governs the field on the *y*-axis and the rotating magnet governs that on the *x*- and *z*-axis. The magnetic flux density in the *x*-axis is lower than the *z*-axis as measuring points at the angle of $\alpha = 90°$ and 270° are the sides of a dipole magnet. The magnetic field vector rotates on the surface of a virtual elliptical cone shape. The shape is not a full circle in the *y-z* plane due to the magnetic field distribution of a permanent magnet. The numerical simulation result of the magnetic flux density in the *xy*-, *yz*-, and *xz*-plane at the center point is visualized in **Figures S2a**, **b**, and **c**, respectively. The red arrows indicate the magnetic field direction at $\alpha = 180°$. All contour lines share the same legend shown in **Figure S2c**. Numerical simulation also shows good agreement with the experimental results (**Figure S2d**).

The analytical solution of the magnetic field is applied to model the dynamics of the TrainBot unit using $\theta$ and $\varphi$, which stand for azimuth and elevation angle, respectively (**Figure 3c**). **Figure 3d** shows the analytical solutions of the angles during a full cycle. The angle $\theta$ changes from 67° to 112° and $\varphi$ changes from -39° to 39° and the maximum difference of each angle was measured as 45° and 78°, respectively. It implies that the TrainBot unit raises its foot with a larger displacement than the in-plane displacement (stride length), which could help the TrainBot to escape from dragging by a mucus layer. **Figure 3e** shows the corresponding moving path of a TrainBot unit and each arrow indicates the moving direction of the foot. The numerical simulation result of the magnetic gradient in *x*-, *y*-, and *z*-direction at the center point are shown in **Figures S2e**, **f**, and **g**, respectively. The magnetic gradient in all directions shows less than 0.06 T/m during a full cycle.

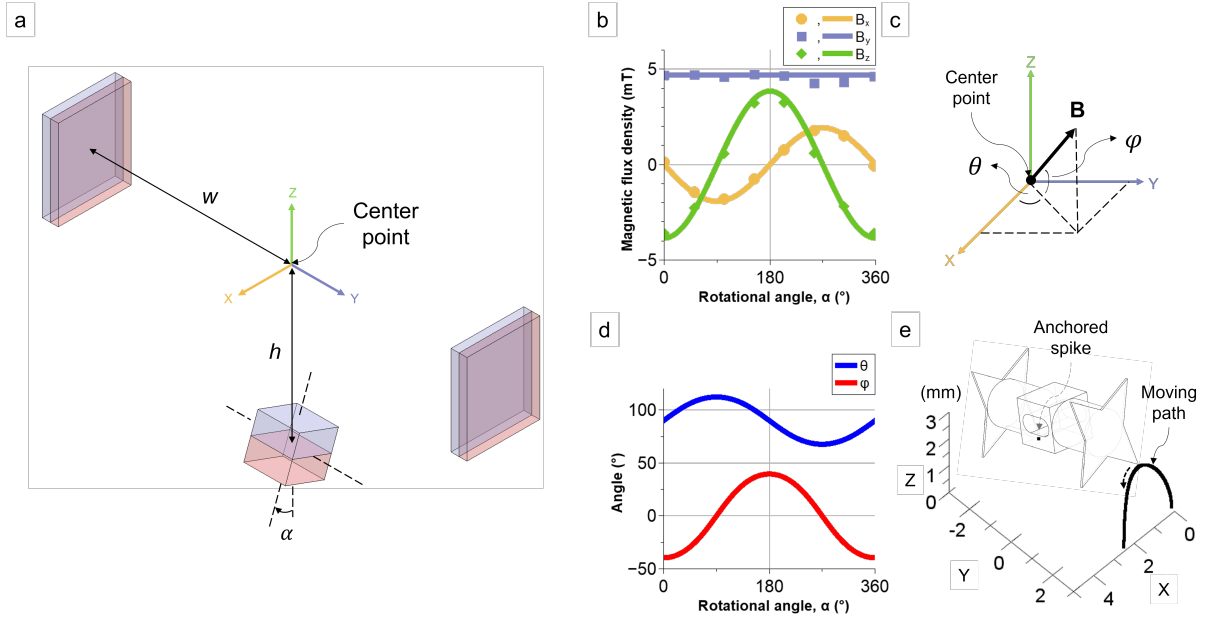

**Figure 3.** Modeling and characterization of the driving magnetic field and TrainBot unit's motion. (a) Schematic of the magnetic actuation set-up (modeled as ideal dipoles) with a distance of *w* and *h* of 259.75 mm and 190.5 mm (center-to-center distance), respectively. During a full rotational cycle (0° ≤ $\alpha$ ≤ 360°), the superimposed magnetic field at the center point is analytically solved, and the magnitudes in three directions are shown in (b). Dots indicate experimental results and solid lines indicate analytical solutions. (c - d) Analytical solution for the azimuth ($\theta$) and elevation ($\varphi$) angle of the external magnetic field **B** during a full cycle. (e) The moving path of the lifting spike of the TrainBot unit is plotted in isometric view at $\alpha$ = 180°, while a spike on the other side is anchored on the ground.

2.3. Characterization of the crawling

The stride length of the TrainBot unit depends on the angle $\varphi$, as shown in **Figure 4a**, which can be varied by changing the height of the rotating magnet *h* and the distance between the static magnet pair *2w*, following **Equation 2-4**. In the experiment, for simple operation, the distance between the two static magnets (center-to-center distance) is kept constant at 519.5 mm, and only the height of the rotating magnet is changed. The experimental result shows excellent agreement with the analytical solution, as plotted in **Figures 4b** and **4c**. It shows that the magnetic flux density $B_z$ decreases from 4.1 to 1.9 mT when the rotating magnet moves from 182 to 238 mm. Accordingly, the TrainBot unit's maximal pose angle ($\varphi_{max}$) decreases from 41° to 22°.

**Figure 4d** illustrates the stride length of a TrainBot unit over a full cycle. **Figure 4e** shows that the measurements are in good agreement with the analytical solution. A longer stride length is observed with a larger pose angle, due to a higher magnetic field strength $B_z$. This result is consistent with the observation of the stride lengths, as each stride measures 2.5 mm at 41°, which is 160% and 220% larger than that of 31° and 22°, respectively. **Figure 4f** shows the frequency response of the TrainBot unit. As the rotational frequency increases from 0.17 Hz to 1.7 Hz, the TrainBot unit's forward velocity increases from 0.4 mm/s to 4.2 mm/s linearly for all three maximal pose angles. At all frequencies, a TrainBot unit with a pose angle at 41° moves 1.6 and 2.2 times faster than that at 31° and 22°, respectively.

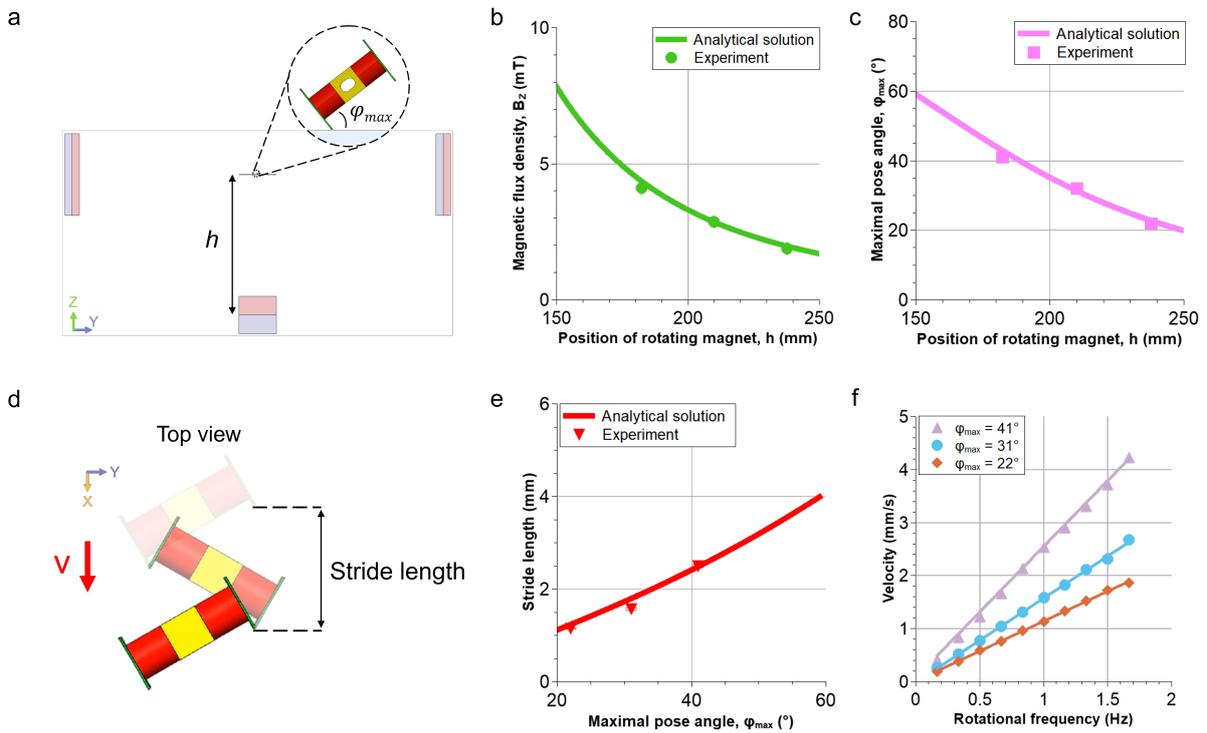

**Figure 4.** Characterization of a TrainBot unit's locomotion. a) Schematic front view of the magnetic actuation set-up and TrainBot unit's maximal pose angle $\varphi_{max}$. b) Magnetic flux density in the *z*-direction at different positions of the rotating magnet. c) TrainBot unit's pose angles at different locations of the rotating magnet. d) Schematic top view of a TrainBot unit's full locomotion cycle. e) Stride lengths at different maximal pose angles. f) Comparison of the crawling speed at different actuation frequencies.

2.4. TrainBot units' feet design and characterization

The magnetic actuation set-up enables the TrainBot unit's rotational motion, but in order to move forward on a slippery surface, the unit needs to exert a sufficient pushing force against the substrate, *i.e.*, the static foot should be firmly anchored to the surface without slipping. For this reason, special spikes (~10 μm at the apex) are designed on the feet to help the TrainBot unit maintain its position by anchoring into the soft biological substrate. **Figure 5a** shows the design of a foot. The foot has four spikes with the design parameters $l_1$, $l_2$, $l_3$, and $\delta$. For optimization of the foot, in **Figure 5b**, the spike angle $\delta$ is varied for three different foot designs ($\delta$ = 15°, 30°, and 45°), while the other three parameters were kept constant ($l_1$ = 0.5 mm, $l_2$ = 0.8 mm and $l_3$ = 1 mm).

**Figure 5c** presents a schematic of the set-up to measure the pulling force of the TrainBot unit. The force was measured when the TrainBot unit was actuated on a soft hydrogel phantom surface by the external magnetic field. As shown in **Figure 5d**, the TrainBot unit without feet generates the smallest force (0.27 mN). When a TrainBot unit has a larger spike ($\delta$ = 45°), the robot exhibits around 3.4 times higher force (0.93 mN) than a robot without feet at the same magnetic torque. Intuitively, a sharp apex can penetrate the slippery but soft substrate to facilitate a firm anchoring point, and thus it increases the forward propulsive force. However, it is surprising that the width of the spike has an opposite impact on the propulsive force, *i.e.,* a wider spike results in a larger propulsive force, as measured in the experiments.

To understand this phenomenon, the penetration depth of the spike into the soft substrate was imaged microscopically to show the side view of the TrainBot unit with different spike angles. As shown in **Figure 5e,** we observed a gap between the TrainBot unit and the substrate (marked in red) with the spike angles at 15° and 30°, which does not appear with the spike angle at 45°. Moreover, we observed that two spikes are in contact with the substrate, but only one spike penetrates the surface. The same observation is found in the geometrical simulation of the TrainBot units, as shown in the second row of **Figure 5e**. An enlarged figure illustrates the penetrating depth (*d*) and area (*A*) of a spike. During the actuation, a foot anchors to a substrate while lifting the other foot. The force induced by the hydrogel phantom while lifting a foot is characterized by a simplified dynamic condition, which results in a lifting force of ~0.1 mN (**Figure S3**). In dynamic conditions, however, the anchored spike rotates while maintaining its position along with the magnetic gradient and body weight, and the maximum magnetic torque (~10 mN ·mm) generates a lifting force of ~3 mN, which is way larger than the lifting force needed for the foot of the robot. **Figure 5f** shows both the simulation and experimental results. The simulation result shows that the maximal penetrating depth (0.84 mm) is achieved with a spike angle of 41°, which aligns well with the experimental data where the spike with an angle of 45° offers the deepest penetration.  The experimental result of the penetrating depth shows good agreement in trend with an average error of 14% lower than the measurements. The main reason is that the deformation of the soft substrate was not considered in the simulation. The direct measurement of the spike area under the surface is inaccurate by experiments, as the spike penetrating the surface is not visible. However, simulation results suggest that the largest penetrating area (0.27 mm$^2$) is achieved with a spike angle of 41°. Considering a finite pressure limit of the biological materials without mechanical fracture, a larger penetrating area of the spike, *i.e.*, a larger contact area between the TrainBot unit and the substrate, leads to a higher maximal propulsive force that can be tolerated by the tissue without fracture. Therefore, the theory explains that 45° is the optimal spike angle for the TrainBot unit's feet design that offers the highest propulsive force.

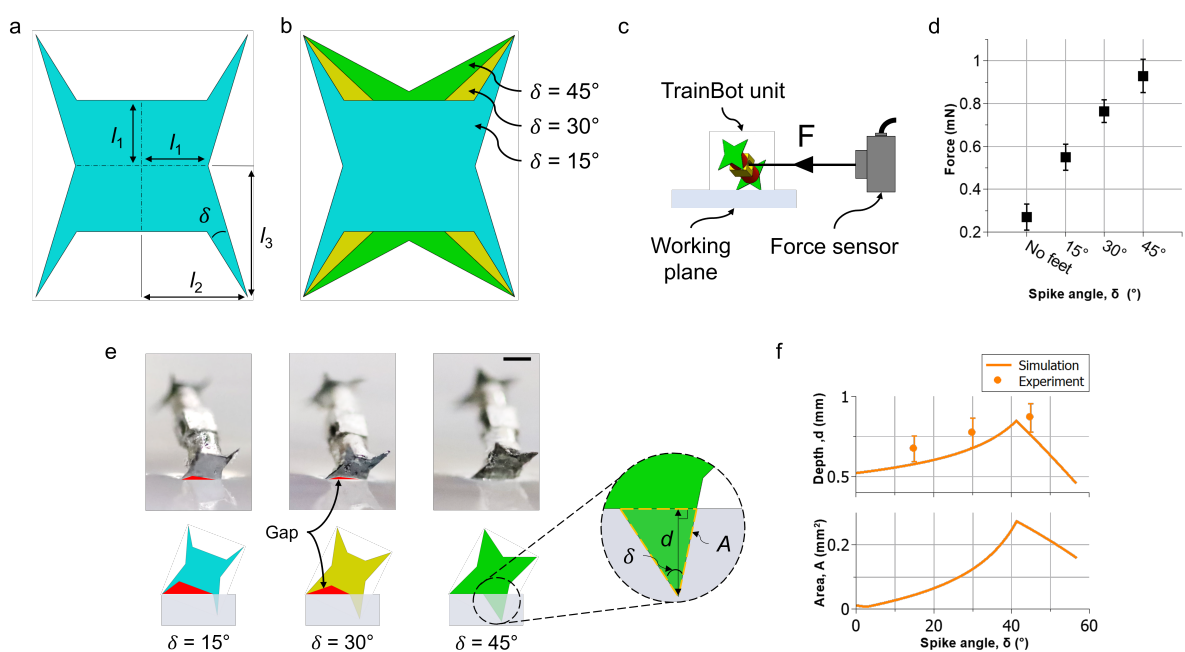

**Figure 5.** Characterization and modeling of the feet design. (a) Design parameters of the feet. A spike angle $\delta$ of 15° is shown as an example ($l_1$ = 0.5 mm, $l_2$ = 0.8 mm and $l_3$ = 1 mm). (b) Three different foot designs ($\delta$ = 15° in turquoise, 30° in yellow, and 45° in green) and images of the fabricated feet are shown in **Figure S4**. (c) Schematic of the pulling force measurement set-up. A TrainBot unit is attached to a force sensor by a wire and actuated on a hydrogel phantom. (d) Force generated by a TrainBot unit with different feet designs, compared to a robot without feet. (e) Comparison of penetration depth for three feet designs on a hydrogel phantom. The first and second rows show the experimental and simulation results, respectively. Enlarged view of a foot penetrating the hydrogel phantom (*d*: penetrating depth, *A*: penetrating area). The scale bar is 1 mm. (f) Experimental- and simulation results of the penetration depth for different feet designs. The penetrating area is obtained by simulation.

## 2.5. Control of a TrainBot unit

The magnetic actuation set-up has one rotational DoFs (Degrees of Freedom), which are useful to control the TrainBot. **Figure 6** and **Video S1** show the actuation of the TrainBot unit on a horizontal plane, on a vertical surface, and an upside-down surface. **Figure 6a** shows a letter 'S' drawn by a TrainBot unit. The unit was actuated by the rotating magnet to move forward. For steering, the angle $\beta$ was manually controlled to follow a predesigned trajectory. Our magnetic actuation set-up enables the TrainBot unit to move not only on a horizontal surface but also on a vertical and an upside-down surface since the unit can be fixed on a surface by the magnetic gradient force against gravity (the direction labeled in **Figure 6**. A higher magnetic gradient is required to keep the TrainBot unit on a vertical surface or an upside-down surface. The minimum magnetic gradient for the current robot (20 mg in weight) on the vertical and the upside-down surfaces are 0.35 T/m and 0.6 T/m respectively. The permanent magnetic actuation set-up can be adapted (details in **Figure S5**) to perform the task, *i.e.,* to be placed vertically to actuate the TrainBot unit on the vertical surface and to be inverted for the upside-down crawling, which shows the versatility of the actuation set-up. Following the same magnetic actuation scheme, the TrainBot unit can also be scaled down to an even smaller scale. As shown in **Figure 6d**, the micro-scale TrainBot consists of three microrobots that are made by laser cutting superparamagnetic material and connecting them by a customized polymeric wire (**Figure S6**). Each microrobot has dimensions of 0.40 × 0.2 × 0.18 $mm^3$, and a group of three microrobots forms a convoy that is 4 mm in length. It shows fully-controllable motion similar to the TrainBot unit and the in-plane control to follow complex trajectory is also shown in **Figure 6e**. It suggests that the reported robot actuation and control mechanism is general and can be readily down-scaled to the sub-millimeter range to enable new minimally-invasive procedures.

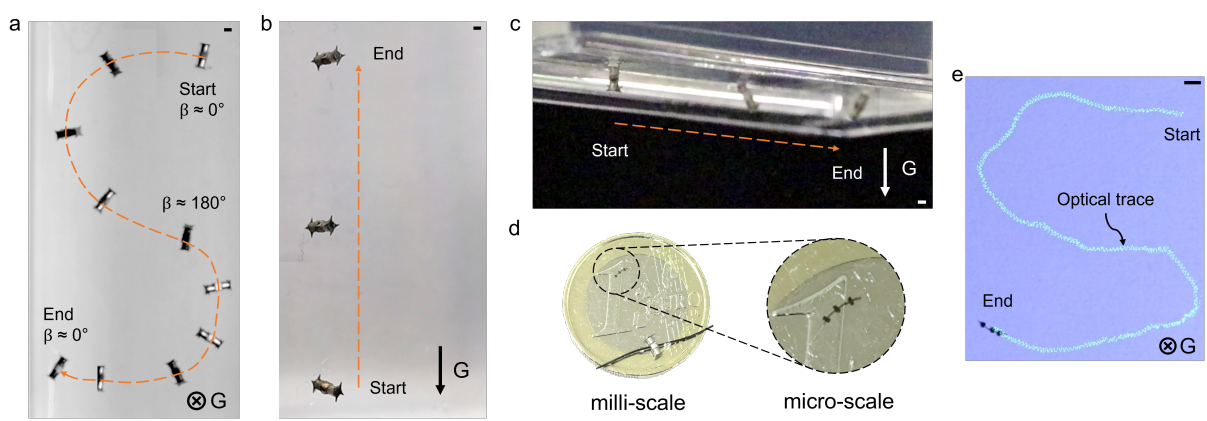

**Figure 6.** Crawling of the TrainBot unit with the optimized feet (a) on a horizontal surface to follow a complex trajectory, (b) on a vertical surface against gravity, (c) upside-down on the bottom of a surface. (d) Photo of the milli- and micro-scale TrainBot on a coin. (e) Micro-scale TrainBot crawls to follow a complex trajectory on a horizontal surface. The symbols labeled with the "G" indicate the gravity direction. All scale bars are 1 mm.

2.6. Biomedical application of TrainBot

In real medical applications, single millirobots often provide a limited force that is too small to perform useful surgical tasks and exhibit a high risk of being stuck in the slippery terrain *in vivo*. We present a new concept by grouping multiple millirobots (TrainBot) in a group to assist one another in effectively crawling on slippery biological surfaces and carrying a long and heavy endoscopic instrument. The schematic design and the real picture of the TrainBot are shown in **Figures 7a** and **7b**. Three TrainBot units are connected by a wire through the central hole of each coupler, and separated with each other by stoppers. The stoppers keep the distance between individual units and also transmit the actuation force by the individual unit to push (or pull) the endoscopic instrument for insertion (or retrieval). Kinematically, the minimum distance between the TrainBot units is 3 mm. To secure working distance and proper motion, each unit is assembled with a distance of ~5 mm.

The group of TrainBot units generates a higher propulsive force than a single unit. As shown in **Figure 7c**, the maximal propulsive force is 1.6 times higher with two units in a group, and 1.8 times higher with three units in a group. The resultant forces do not scale proportional to the number of units due to the phase lag among different units actuated by the magnetic field distributed over the distance. Numerical simulation shows the maximum phase lag between Trainbot units is 24° (details in **Figure S7**). Therefore, the maximum force is not proportional to the number of TrainBot units. There are diminishing returns for each additional robot because they cannot be too close to each other. However, the asynchronous phase of robots may result in a more continuous forward force, which means more Trainbot units could generate a higher mean pulling force, as the units work sequentially.

It is observed in the experiments using *ex vivo* porcine intestine and bile duct that the group of three TrainBot units shows stable locomotion and a high capability to overcome obstacles. **Figure 7d** shows an image sequence of the TrainBot crawling over a fold on the intestinal surface (~2 mm height, see **Video S3**). At the time point $t_2$, the 1$^{st}$ unit failed to anchor on the tissue due to the height of the fold, but the 2$^{nd}$ and 3$^{rd}$ units were able to keep the anchoring and continue to pull the electrical wire, as shown in the time frame $t_3$. Following the same method, the 2nd and 3rd units were able to crawl over the fold (at the time frames $t_4$ and $t_5$).

An important feature of the TrainBot is that it can carry a load that is much heavier and longer than the TrainBot unit itself. For example, we demonstrate an electrical wire (250 mm in length, ~150 times of the TrainBot unit; and 70 mg in weight, 3.5 times of the TrainBot unit) can be carried to the target location for electrocautery. **Figure 7e** shows the schematic of the set-up for the electrocautery experiment. A commercial endoscope records the video from the back side of the obstacle. The TrainBot is inserted into the bile duct, which has a slippery surface and is covered with slippery mucus. The TrainBot can effectively crawl in the bile duct and carry an electrical wire, as shown in **Figure 7f**. The common bile duct where the TrainBot is designed to actuate is the longest and thickest segment of the bile duct.[44] The spikes on the foot of a TrainBot can be optimized to minimize the damage to the biological tissues, as the mucosa layer thickness varies on different organs, *e.g.* the mucus thickness ranges from 10 to 750 μm in small and large intestines in humans.[21,44–46] Electric current is applied through the wire and the cauterization was carried out at the exposed metallic tip (**Figure 7g**). By controlling the orientation of TrainBot, the tissue that blocks the duct is gradually resected. By repeating the resection procedure, a large enough opening (marked using the red dashed line at $t_4$) was achieved. Afterward, the electric wire is retrieved by reversing the motion of the TrainBot, and a soft catheter is transported by another TrainBot passing through the opening for fluid drainage or local drug delivery (shown in a human-scale transparent bile duct phantom in **Figure S8**). The experiment demonstrates the first electrocautery procedure performed by a group of TrainBot units. It shows that multiple units can cooperate to carry surgical instruments that are much longer and heavier than a single TrainBot unit. The active locomotion of the TrainBot units enables the dragging of the instruments along a narrow biological lumen, which outperforms the traditional pushing method of endoscopic instruments by offering higher dexterity and better safety for navigating complex physiological structures.

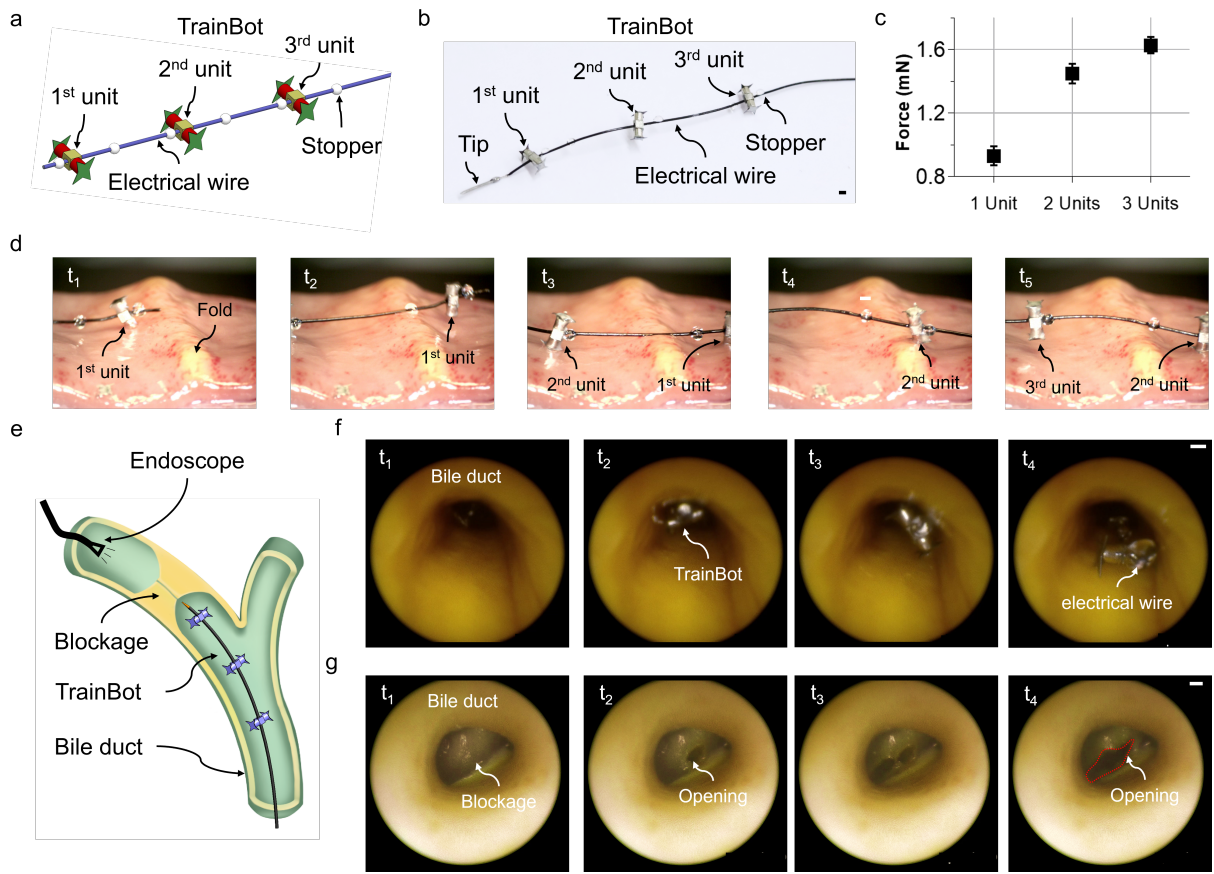

**Figure 7.** TrainBot for endoscopic electrocauterization. (a) Schematic of the TrainBot. Three TrainBot units are connected by a wire and confined with stoppers. (b) Picture of the TrainBot. The scale bar is 1 mm. (c) Force generated by different numbers of TrainBot units. (d) TrainBot crawling on an intestine surface over a circular fold (~2 mm in height, **Video S3**). (e) Schematic showing the experimental set-up for an *ex vivo* electrocautery experiment. An endoscope is inserted from the other end of the porcine bile duct to observe the motion of the TrainBot. The endoscope is positioned behind the blockage for observation. (f) Endoscopic view of the TrainBot crawling in the bile duct to carry an electrical wire. (g) Endoscopic view of the TrainBot performing an electrocauterization to resect the blocked tissue and make an opening.

3. Conclusion

In this work, we report the TrainBot – a group of cooperative TrainBot units that form a convoy to crawl on biological tissue surfaces and carry long endoscopic instruments. By optimizing the anchoring feet's spike angle, an increase of the propulsive force on a slippery surface of factor three could be achieved. A human-scale magnetic set-up is developed to wirelessly actuate and control the robots, which not only allows the precise control of in-plane locomotion but also enables the crawling of the TrainBot unit on vertical and upside-down surfaces. The TrainBot effectively crawls in the *ex vivo* intestine and bile duct and carries endoscopic instruments to successfully perform an electrocautery procedure. The new cooperative strategy of using distributed multiple TrainBot units enables the transportation of an instrument that is much longer and heavier than a single TrainBot unit, which may lead to new endoscopic procedures for useful medical applications.

4. Experimental Section

Design and Fabrication of the TrainBot

As shown in **Figure 2a**, a TrainBot unit is comprised of three components: two actuators, a coupler, and two feet with spikes. The TrainBot unit was designed in SOLIDWORKS 2021 (Dassault Systemes, France). The actuator is a cylindrical permanent magnet magnetized along the longitudinal direction (NdFeB-N45, 1 mm outer diameter, 1 mm in height, Supermagnete, Germany). The coupler is made of mu-metal (>99% purity, 1 mm thick) fabricated by femtosecond laser cutting (MPS FLEXIBLE, ROFINBAASEL Lasertech, Germany). The foot is made of Molybdenum (>99% purity, 50 μm thick) with four spikes (~10 μm at the apex), fabricated by femtosecond laser cutting (MPS FLEXIBLE). The fabricated components are shown in **Figure S4**. These components were assembled using an adhesive (Loctite 401, Henkel, Germany). Two actuators are attached on each side of the coupler, and then the feet with spikes are attached to both ends of the actuators. Multiple TrainBot units are connected by an electrical wire (OD 0.35 mm) through the central hole of each coupler. Each TrainBot unit was secured by the stoppers using epoxy glue (UHU PLUS, UHU GmbH, Germany) on the wire to prevent magnetic attraction between TrainBot units (**Figure 7b**). The sub-millimeter scale TrainBot (**Figure S7**) was made by laser cutting mu-metal material to a block of 0.4× 0.20 × 0.18 mm$^3$ with a central hole of 0.1 mm in diameter. A central wire was customized by laser cutting a thin polymer film (50 μm thick) and assembled through the central hole of three mu-metal blocks under a stereoscope (OZP 552, KERN & SOHN GmbH, Germany).

Permanent Magnetic Actuation Set-up

The TrainBot is actuated with a customized magnetic actuation set-up shown in **Figure 2a**. The set-up comprises three permanent magnets: a pair of static magnets and a rotating magnet. The static magnets (NdFeB-N45, 110.6 × 89 × 19.5 mm$^3$, Supermagnete) were fixed with a gap of 519.5 mm (center to center distance) that generates a static magnetic field along the *y*-axis. The rotating magnet (NdFeB-N40, 50.8 × 50.8 × 50.8 mm$^3$, Supermagnete) was placed at the center of the two static magnets in the *y*-direction at a distance of *w*, and the distance *h* in the *z*-direction can be adjusted in the range of ~110-250 mm (details in **Figure S1b**). To characterize the superimposed magnetic field, the magnetic moment of each magnet was calculated by **m** = **M**V. Here, M is the magnetization per unit volume, and V is the volume of the magnet.[47] The magnetization is calculated using M = $B_r/\mu_0$, where $B_r$ is the remanence and $\mu_0$ is the vacuum permeability.[48] The median remanence values specified in the datasheets of the magnets[49], *i.e.* 1.345 T for N45 (static magnet) and 1.275 T for N40 (rotating magnet), were used in the calculation, which results in the $m_s$ and $m_r$ of 205.4 A·m$^2$ and 133.0 A·m$^2$, respectively. It generates a rotating magnetic field by controlling the angle *α* using a stepper motor (23HS30-2804Sm, OMC Co. Ltd., China) and a microcontroller (Arduino UNO, Italy). The set-up achieves 5~6 mT depending on the angle *α* in an accessible volume of ⌀ 50 × 28 cm$^2$. The resulting magnetic flux density was measured by a magnetometer (Hall Probe C with F3A Magnetic Field Transducers, SENIS, Switzerland) with a motorized scanning system (SF600, GAMPT, Germany). Numerical simulation of the magnetic flux density was performed using the AC/DC Module in COMSOL Multiphysics (v5.6, COMSOL AB, Stockholm, Sweden).

The magnetic actuation set-up achieves two DoF as control inputs by controlling angle *α* and *β* (**Figure 2a**). The TrainBot unit's translational motion was controlled by varying the angle *α*, and its rotational motion, *i.e.,* the unit's moving direction, was controlled by rotating the magnetic actuation set-up around the *z*-axis to the angle *β*.

Preparation of Animal Tissues

To mimic real clinical conditions, *ex vivo* experiments were performed with fresh animal tissues, including fresh porcine intestines, porcine stomachs, porcine bile ducts, and chicken breasts. These tissues were all obtained from a local butcher. The porcine intestines were dissected from the upper part of the jejunum. The porcine stomachs were dissected from the body part. The bile ducts were obtained from porta hepatis. Both ends of the bile duct were sutured and fixed by a structure to keep the internal shape using a customized 3D printed part (Clear Resin, Form3L, Formlabs, USA). A small piece of chicken breast was dissected and sutured in the bile duct to mimic biliary stenosis. The animal tissues were stored at 4 °C and used for experiments within 24 h after the sacrifice of the animal.

Electrocautery test using TrainBot

A TrainBot assembled using an electrical wire (OD 0.35 mm) is connected to an electrocautery device (HF-Generator 2260, Richard Wolf GmbH, Germany). For the electrocautery test, a monopolar was used while being grounded on the animal tissue kept wet using a saline solution (0.9%). A rigid endoscope (8703.523, Richard Wolf GmbH) connected to a camera (5521902 and ENDOCAM Flex HD, Richard Wolf GmbH) was positioned behind the blockage shown in **Figure 7e** to observe the crawling and resection in a bile duct.

Human-scale bile duct phantom

The bile duct phantom was built using the SOLIDWORKS 2021 (Dassault Systemes) and Meshmixer (Autodesk Meshmixer v3.5, CA, USA). The biliary duct structure was printed using a soft photopolymer (Elastic 50A Resin) on a 3D printer (Form3L, Formlabs).[50]

Characterization of the Crawling Motion

The TrainBot unit was tested on a hydrogel phantom and animal tissues to characterize its crawling capability. The hydrogel phantom was prepared using gelatin from porcine skin (3.4%wt, G1890, Sigma-Aldrich, Germany). The gelatin powder was dissolved in deionized water at 65 °C for 1 h and then cooled down to be stored at 4 °C before testing. All experiments were carried out at room temperature (22 °C).

The locomotion was imaged using a Canon EOS RP (Canon, Japan) and a high-speed camera (Kron Technologies, Canada) with a 60 mm lens (Canon). The video was analyzed using an image analysis tool (ImageJ, 1.53t, National Institutes of Health, USA).

Simulation of the TrainBot unit's crawling was performed in SOLIDWORKS 2021 (Dassault Systemes) using a module of motion study. The motion path was analyzed with analytical solutions.

Force Measurement Set-up

TrainBot units with three different spike angles (**Figure 5b**) as well as without feet were tested on a hydrogel phantom. Force measurements were performed using a force sensor (403A-5mN, Aurora Scientific Inc, Canada) attached through the coupler by a thread (Guetermann GmbH, Germany). Each type of TrainBot unit was measured at least five times, and its peak value was used to calculate an average value. The acquired data was postprocessed using customized codes in MATLAB 2021b (MathWorks Inc., MA, USA).

Lifting Force Measurement Set-up

Three different designs of foot (**Figure 5b**) were tested on a hydrogel phantom. Each foot design was attached to the Force sensor (403A-5mN) using the thread (Guetermann GmbH). One spike was placed in the phantom and pulled out in the vertical direction (**Figure S3**). Measurement was performed at least five times for each foot design, and the data was post-processed using customized codes in MATLAB 2021b (MathWorks Inc.).

**Supporting Information**

Supporting Information is available from the Wiley Online Library or from the author.

**Video S1**: The crawling of a TrainBot unit

**Video S2**: Human-scale magnetic actuation set-up

**Video S3**: TrainBot crawling on biological tissues

Acknowledgments

This work was partially funded by the European Union (ERC, VIBEBOT, 101041975), the Vector Foundation (Cyber Valley research group), and the MWK-BW (Az. 33-7542.2-9-47.10/42/2). M. J. and T. Q. acknowledge the support of the Stuttgart Center for Simulation Science (SimTech).

Received: ((will be filled in by the editorial staff))

Revised: ((will be filled in by the editorial staff))

Published online: ((will be filled in by the editorial staff))

References

[1] B. Wang, K. Kostarelos, B. J. Nelson, L. Zhang, *Advanced Materials* **2021**, *33*, 2002047.
[2] C. K. Schmidt, M. Medina-Sánchez, R. J. Edmondson, O. G. Schmidt, *Nat Commun* **2020**, *11*, 5618.
[3] J. Li, B. E. F. D. Ávila, W. Gao, L. Zhang, J. Wang, *Science Robotics* **2017**, *2*, 6431.
[4] Z. Wu, J. Troll, H. H. Jeong, Q. Wei, M. Stang, F. Ziemssen, Z. Wang, M. Dong, S. Schnichels, T. Qiu, P. Fischer, *Science Advances* **2018**, *4*, eaat4388.
[5] Y. Su, T. Qiu, W. Song, X. Han, M. Sun, Z. Wang, H. Xie, M. Dong, M. Chen, *Advanced Science* **2021**, *8*, 2003177.
[6] H. Xie, X. Fan, M. Sun, Z. Lin, Q. He, L. Sun, *IEEE/ASME Transactions on Mechatronics* **2019**, *24*, 902.
[7] Z. Yang, L. Yang, M. Zhang, Q. Wang, S. C. H. Yu, L. Zhang, *IEEE Robotics and Automation Letters* **2021**, *6*, 1280.
[8] C. Zhou, Y. Yang, J. Wang, Q. Wu, Z. Gu, Y. Zhou, X. Liu, Y. Yang, H. Tang, Q. Ling, L. Wang, J. Zang, *Nat Commun* **2021**, *12*, 5072.
[9] J. M. Prendergast, G. A. Formosa, M. E. Rentschler, *IEEE Robotics and Automation Letters* **2018**, *3*, 2670.
[10] G. A. Formosa, J. M. Prendergast, S. A. Edmundowicz, M. E. Rentschler, *IEEE Transactions on Robotics* **2020**, *36*, 545.

[11] J. Leal, H. D. C. Smyth, D. Ghosh, *Int J Pharm* **2017**, *532*, 555.
[12] S. J. Wu, X. Zhao, *Chem. Rev.* **2023**, *123*, 14084.
[13] Y. Wu, X. Dong, J. Kim, C. Wang, M. Sitti, *Science Advances* **2022**, *8*, eabn3431.
[14] D. Guo, L. Wermers, K. R. Oldham, *IEEE/ASME Transactions on Mechatronics* **2022**, *27*, 1910.
[15] R. Tan, X. Yang, H. Lu, L. Yang, T. Zhang, J. Miao, Y. Feng, Y. Shen, *Matter* **2022**, *5*, 1277.
[16] H. Lu, M. Zhang, Y. Yang, Q. Huang, T. Fukuda, Z. Wang, Y. Shen, *Nat Commun* **2018**, *9*, 3944.
[17] M. Jeong, X. Tan, F. Fischer, T. Qiu, *Micromachines* **2023**, *14*, 1439.
[18] W. Hu, G. Z. Lum, M. Mastrangeli, M. Sitti, *Nature* **2018**, *554*, 81.
[19] J. Li, H. Wang, Q. Shi, Z. Zheng, J. Cui, T. Sun, P. Ferraro, Q. Huang, T. Fukuda, *IEEE Transactions on Nanotechnology* **2020**, *19*, 21.
[20] T. Xu, J. Zhang, M. Salehizadeh, O. Onaizah, E. Diller, *Science Robotics* **2019**, *4*, eaav4494.
[21] C. Atuma, V. Strugala, A. Allen, L. Holm, *American Journal of Physiology-Gastrointestinal and Liver Physiology* **2001**, *280*, G922.
[22] J. J. Abbott, K. E. Peyer, M. C. Lagomarsino, L. Zhang, L. Dong, I. K. Kaliakatsos, B. J. Nelson, *The International Journal of Robotics Research* **2009**, *28*, 1434.
[23] M. H. D. Ansari, V. Iacovacci, S. Pane, M. Ourak, G. Borghesan, I. Tamadon, E. Vander Poorten, A. Menciassi, *Advanced Functional Materials* **2023**, *33*, 2211918.
[24] A. Ghosh, P. Fischer, *Nano Letters* **2009**, *9*, 2243.
[25] X. Wang, C. Ho, Y. Tsatskis, J. Law, Z. Zhang, M. Zhu, C. Dai, F. Wang, M. Tan, S. Hopyan, H. McNeill, Y. Sun, *Science Robotics* **2019**, *4*, eaav6180.
[26] H.-W. Huang, M. S. Sakar, A. J. Petruska, S. Pané, B. J. Nelson, *Nat Commun* **2016**, *7*, 12263.
[27] B. Wang, K. F. Chan, K. Yuan, Q. Wang, X. Xia, L. Yang, H. Ko, Y.-X. J. Wang, J. J. Y. Sung, P. W. Y. Chiu, L. Zhang, *Science Robotics* **2021**, *6*, eabd2813.
[28] F. Fischer, C. Gletter, M. Jeong, T. Qiu, *npj Robot* **2024**, *2*, 1.
[29] Y. Dong, L. Wang, V. Iacovacci, X. Wang, L. Zhang, B. J. Nelson, *Matter* **2022**, *5*, 77.
[30] T. Qiu, M. Jeong, R. Goyal, V. M. Kadiri, J. Sachs, P. Fischer, In *Field-Driven Micro and Nanorobots for Biology and Medicine* (Eds.: Sun, Y.; Wang, X.; Yu, J.), Springer International Publishing, Cham, **2022**, pp. 389–411.
[31] P. Ryan, E. Diller, In *Proceedings - IEEE International Conference on Robotics and Automation*, Institute of Electrical and Electronics Engineers Inc., **2016**, pp. 1731–1736.
[32] D. Son, M. C. Ugurlu, M. Sitti, *Science Advances* **2021**, *7*, eabi8932.
[33] D. Li, M. Jeong, E. Oren, T. Yu, T. Qiu, *Robotics* **2019**, *8*, 87.
[34] M. Jeong, M. Zhang, F. Fischer, T. Qiu, In *International Conference on Manipulation, Automation and Robotics at Small Scales (MARSS)*, Abu Dhabi, United Arab Emirates, **2023**.
[35] A. W. Mahoney, J. J. Abbott, *The International Journal of Robotics Research* **2016**, *35*, 129.
[36] A. A. Connor, S. Kodali, M. Abdelrahim, M. M. Javle, E. W. Brombosz, R. M. Ghobrial, *Frontiers in Oncology* **2022**, *12*.
[37] T. D. Ellington, B. Momin, R. J. Wilson, S. J. Henley, M. Wu, A. B. Ryerson, *Cancer Epidemiol Biomarkers Prev* **2021**, *30*, 1607.
[38] D. I. Sucandy, *Biliary Obstruction: Diagnosis and Treatment Procedure*, **2022**.
[39] R. J. Huang, N. C. Thosani, M. T. Barakat, A. Choudhary, A. Mithal, G. Singh, S. Sethi, S. Banerjee, *Gastrointestinal Endoscopy* **2017**, *86*, 319.
[40] S. Peter, J. Y. Bang, S. Varadarajulu, *Current Opinion in Gastroenterology* **2021**, *37*, 416.
[41] A. Vezakis, G. Fragulidis, A. Polydorou, *World Journal of Gastrointestinal Endoscopy* **2015**, *7*, 1135.
[42] M. A. Khashab, A. Tariq, U. Tariq, K. Kim, L. Ponor, A. M. Lennon, M. I. Canto, A. Gurakar, Q. Yu, K. Dunbar, S. Hutfless, A. N. Kalloo, V. K. Singh, *Clinical Gastroenterology and Hepatology* **2012**, *10*, 1157.

[43] T. L. Chow, *Introduction to Electromagnetic Theory: A Modern Perspective*, Jones and Bartlett Publishers, **2006**.
[44] H. Jackowiak, A. Lametschwandtner, *The Anatomical Record Part A: Discoveries in Molecular, Cellular, and Evolutionary Biology* **2005**, *286A*, 974.
[45] I. STEFANOV, *Turkish Journal of Veterinary & Animal Sciences* **2021**, *45*, 113.
[46] J. Y. Lock, T. L. Carlson, R. L. Carrier, *Advanced Drug Delivery Reviews* **2018**, *124*, 34.
[47] P. B. Visscher, *Fields and Electrodynamics*, John Wiley & Sons Australia, Limited, **1988**.
[48] H. C. Burch, A. Garraud, M. F. Mitchell, R. C. Moore, D. P. Arnold, *IEEE Transactions on Antennas and Propagation* **2018**, *66*, 6265.
[49] Physical Magnet Data, www.supermagnete.de/eng/physical-magnet-data. Accessed 12 June 2024.
[50] X. Tan, D. Li, M. Jeong, T. Yu, Z. Ma, S. Afat, K.-E. Grund, T. Qiu, *Ann Biomed Eng* **2021**, *49*, 2139.